\documentclass[times,twocolumn,final]{elsarticle}
\pdfoutput=1

\usepackage{framed,multirow}

\usepackage{amssymb}
\usepackage{latexsym}

\usepackage{url}
\usepackage{xcolor}
\definecolor{newcolor}{rgb}{.8,.349,.1}

\usepackage{hyperref}
\usepackage{makecell}
\usepackage[switch,pagewise]{lineno} 
\usepackage{times}
\usepackage{epsfig}
\usepackage{graphicx}
\usepackage{amsmath}
\usepackage{bm}
\usepackage{graphicx}
\usepackage{caption}
\usepackage{lipsum}
\usepackage{pifont}

\begin{document}


\begin{frontmatter}

\title{3D-SSGAN: Lifting 2D Semantics for \\ 3D-Aware Compositional Portrait Synthesis}

\author[1]{Ruiqi Liu\fnref{fn1}}

\cortext[cor1]{Corresponding author}
\emailauthor{liurq21@mails.jlu.edu.cn}{Ruiqi Liu}
\emailauthor{zhengpeng22@mails.jlu.edu.cn}{Peng Zheng}
\emailauthor{yewang22@mails.jlu.edu.cn}{Ye Wang}
\emailauthor{ruim@jlu.edu.cn}{Rui Ma}
\author[1]{Peng Zheng\fnref{fn1}}

\author[1]{Ye Wang }
\author[1,2]{Rui Ma \corref{cor1}}
\fntext[fn1]{Co-first authors.} 
\address[1]{School of Artificial Intelligence, Jilin University, Changchun, 130012, China}
\address[2]{Engineering Research Center of Knowledge-Driven Human-Machine Intelligence, MOE, China}

\begin{abstract}

Existing 3D-aware portrait synthesis methods can generate impressive high-quality images while preserving strong 3D consistency.
However, most of them cannot support the fine-grained part-level control over synthesized images.
Conversely, some GAN-based 2D portrait synthesis methods can achieve clear disentanglement of facial regions, but they cannot preserve view consistency due to a lack of 3D modeling abilities.
To address these issues, we propose 3D-SSGAN, a novel framework for 3D-aware compositional portrait image synthesis.
First, a simple yet effective depth-guided 2D-to-3D lifting module maps the generated 2D part features and semantics to 3D.
Then, a volume renderer with a novel 3D-aware semantic mask renderer is utilized to produce the composed face features and corresponding masks.
The whole framework is trained end-to-end by discriminating between real and synthesized 2D images and their semantic masks.
Quantitative and qualitative evaluations demonstrate the superiority of 3D-SSGAN in controllable part-level synthesis while preserving 3D view consistency.
\end{abstract}


\end{frontmatter}


\section{Introduction}

\begin{figure*}
    \centering
    \includegraphics[width=\textwidth]{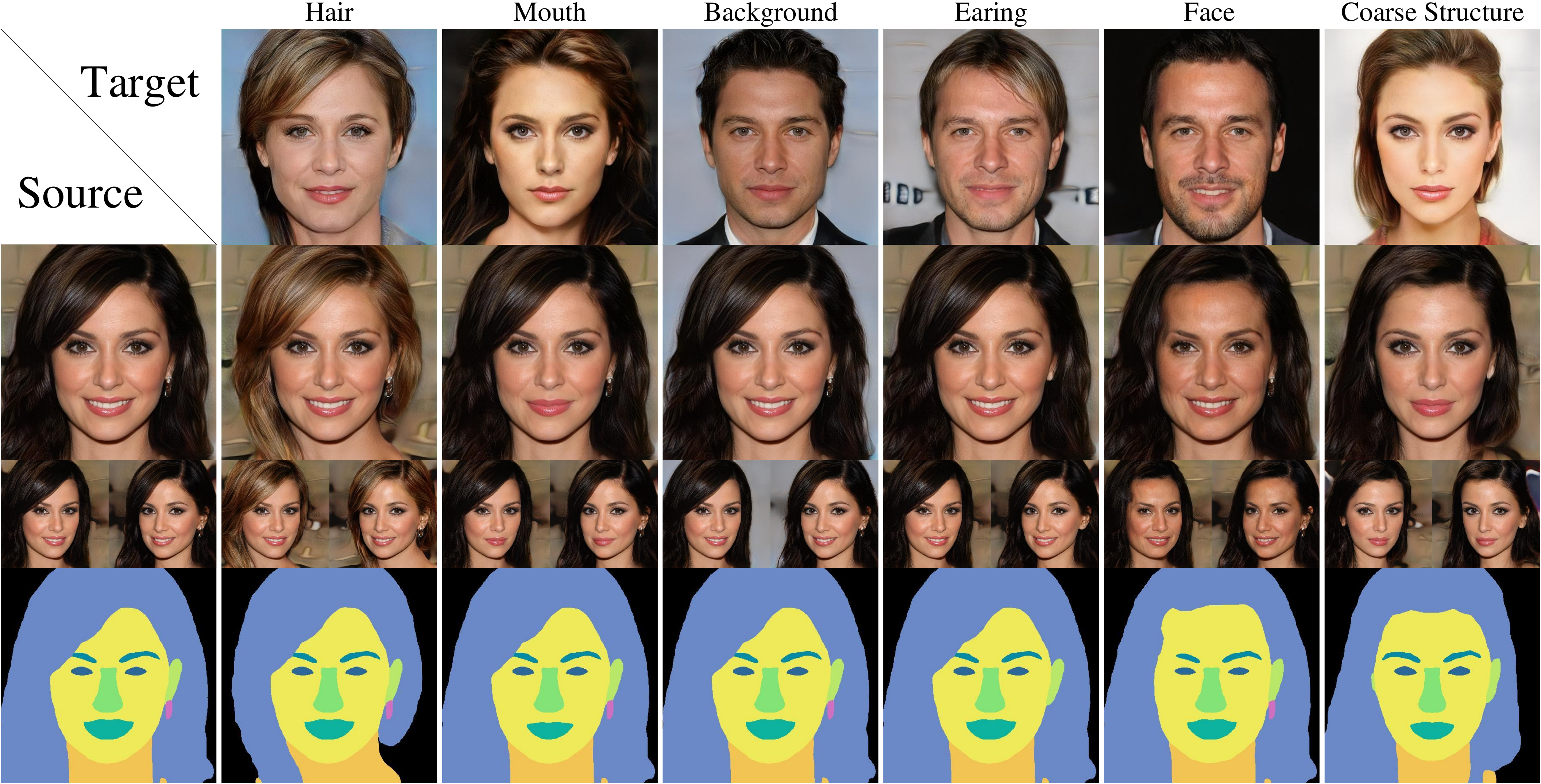}
    \caption{3D-aware part-level synthesis results of 3D-SSGAN.
    Semantic parts in target images are transferred on source image by using the same latent code that generates the target part.
    The images generated from other views and the synthesized masks are shown in the last two rows. }
    \label{fig:teaser}
\end{figure*}
Portrait synthesis and editing have attracted wide attention recently.
Methods like StyleGAN \cite{karras2019style} and its follow-ups \cite{karras2020analyzing,karras2021alias,bermano2022} can generate visually appealing portrait images through global-level modeling.
Furthermore, fine-grained part-disentangled portrait generation is also enabled by explicitly modeling the 2D semantic parts and synthesizing the image in a compositional way such as in SemanticStyleGAN (SSGAN) \cite{shi2022semanticstylegan}.
However, these 2D GAN-based methods are limited in their ability to produce 3D view-consistent portraits as the generation is mainly considered in the 2D domain.
On the other hand, a series of works \cite{gu2021stylenerf,xu20223d,chan2022efficient,deng2022gram,or2022stylesdf} attempt to achieve 3D-aware portrait synthesis.
3D GAN \cite{goodfellow2014generative} and NeRF \cite{mildenhall2021nerf} are employed in these methods to generate multi-view portrait images while maintaining view consistency.
A common practice is to first learn a 3D feature volume or tri-plane representation and then apply volume rendering to generate a view-conditioned portrait image from a rendered 2D feature map.
While these 3D-aware models can synthesize high-quality and view-consistent portraits, they cannot enable editing at the part level.

To achieve both part-level editing and 3D-aware synthesis, approaches such as IDE-3D \cite{sun2022ide} and NeRFFaceEditing \cite{jiang2022nerffaceediting} incorporate semantic masks to the generative learning process and manipulate the masks to edit the images synthesized by 3D neural renders.
Since the images and masks are generated from the latent codes in a holistic manner, the synthesis and editing results of \cite{sun2022ide,jiang2022nerffaceediting} are not strongly disentangled.
For example, for IDE-3D and NeRFFaceEditing, when manipulating a person's mouth to create a smile expression, regions other than the mouth area may be unintentionally changed.
One promising solution to address above issues is CNeRF \cite{ma2023semantic}, which learns an independent 3D generator for each semantic part and performs volume aggregation and rendering to obtain the composed image.
However, as training multiple 3D generators is memory-intensive and potentially unstable, their final synthesis results may be constrained by the training performance.

In this paper, we propose 3D-SSGAN, a novel framework for 3D-aware compositional portrait image synthesis.
Our work is inspired by SSGAN \cite{shi2022semanticstylegan} which explicitly generates the semantic parts from independent latent codes to ensure strong semantic disentanglement.
Also, we incorporate the NeRF-based neural rendering architecture from 3D-aware synthesis methods such as VolumeGAN \cite{xu20223d} to achieve view-consistent generation.
Our goal is similar to CNeRF \cite{ma2023semantic}, but we design two novel modules to reduce the training efforts and attain more semantic-disentangled composition.
Firstly, instead of training 3D generators for the semantic parts, we train 2D generators and apply a simple yet effective depth-guided 2D-to-3D lifting module to directly map the generated 2D part features and semantics to 3D.
It is worth noting that the initial 3D features and semantics lifted from 2D will be further optimized when training the volume renderer.
Hence, we achieve efficient 3D feature generation via simple lifting rather than 3D generators as in CNeRF.
Secondly, we enhance the volume renderer with a novel 3D-aware semantic mask renderer.
This module takes the lifted part semantics (denoted as part density) as input and integrates the learned 3D information to the mask rendering process.
As a result, the generated semantic masks are more plausible for non-frontal views.
The whole framework of 3D-SSGAN is trained end-to-end by discriminating between real and synthesized 2D images and their semantic masks.
Fig.~\ref{fig:teaser} shows the results of our 3D-aware part-level synthesis framework.
By explicitly generating the semantic parts and performing the synthesis using the lifted and composed 3D features, we can enable semantic-disentangled and 3D-consistent portrait synthesis. 

We conduct quantitative and qualitative experiments to compare 3D-SSGAN with state-of-the-art prior works.
Results demonstrate the superiority of 3D-SSGAN in controllable part-level synthesis while preserving 3D view consistency.
Ablation studies are also conducted to verify the effectiveness of our key modules.
In summary, our main contributions are as followings:

\begin{itemize}
    \item We propose 3D-SSGAN, a novel framework for simultaneous strong-disentangled part-level generation and 3D-aware portrait synthesis.
    \item We propose a simple yet effective depth-guided 2D-to-3D lifting module to obtain 3D part features and semantics from 2D generators, avoiding training cumbersome 3D generators.
    \item We propose a novel 3D-aware semantic mask renderer that integrates the learned 3D information for rendering view-consistent semantic masks. 
\end{itemize}

\section{Related Work}
\begin{figure*}
    \centering
    \includegraphics[width=\textwidth]{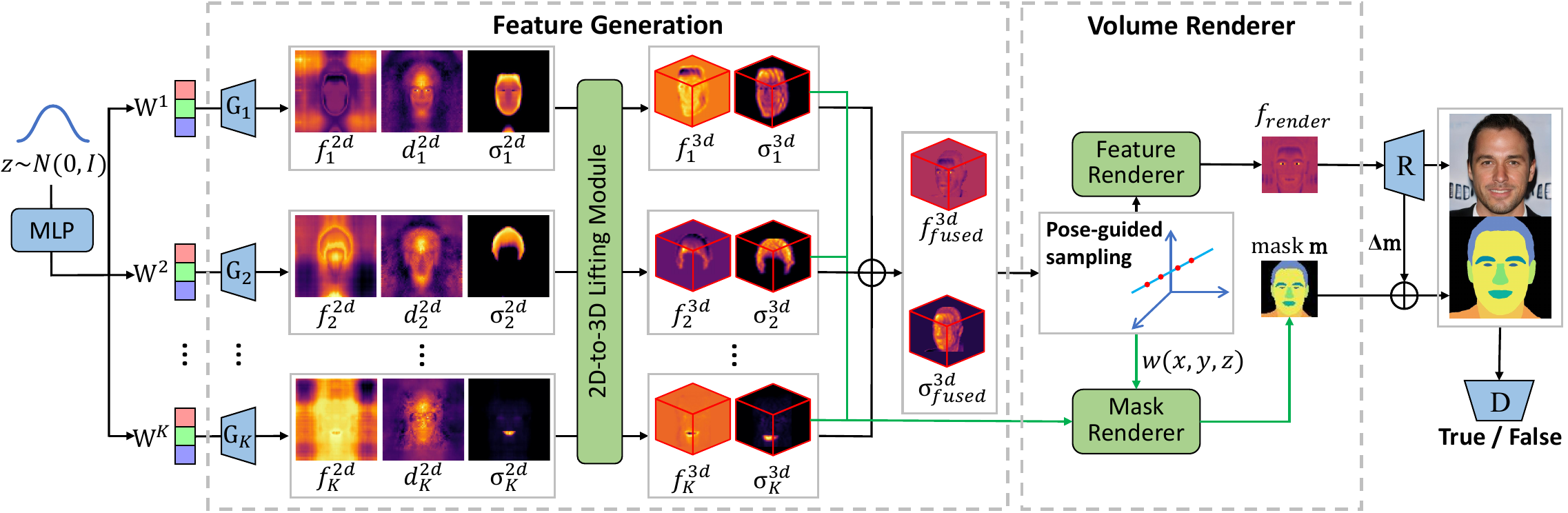}
    \caption{Overview of our training pipeline.
    First, a depth-guided 2D-to-3D lifting module converts 2D part feature $f_k^{2d}$ and density $\sigma_k^{2d}$ generated by independent 2D generators to 3D.
    Then, a posed 2D feature $f_{render}$ and an initial mask $m$ are synthesized by a volume renderer which contains a 3D-aware semantic mask renderer.
    Finally, high-resolution images and masks generated by a render net $R$ are passed to a discriminator $D$ for adversarial training. The green arrows show the data flow for the 3D-aware mask renderer.
    }
    \label{pipeline}
\end{figure*}

Image synthesis has made rapid progress in recent years \cite{bermano2022,tewari2022,xia2022survey}.
In this section, we mainly review recent GAN-based portrait image synthesis and editing works, with the focus on methods which consider 3D consistency or semantics during the synthesis.

\textbf{3D-aware portrait images synthesis.}
Comparing to 2D portrait images synthesis which mainly focus on synthesizing high-resolution and photo-realistic 2D portrait images \cite{karras2019style,karras2020analyzing,karras2021alias}, 3D-aware portrait image synthesis aims to generate 3D-consistent face images across different views.
Early approaches use voxel-based GANs \cite{zhu2018visual,henzler2019escaping,nguyen2019hologan,nguyen2020blockgan} to learn low-resolution 3D features which are projected to 2D for image generation. 
However, these methods struggle to synthesize complex scenes or fine facial details due to the constrained resolution.
Recent 3D-aware synthesis methods \cite{chan2021pi,pan2021shading} combine GAN with NeRF \cite{mildenhall2021nerf} which has shown remarkable results for synthesizing realistic novel view images.
Moreover, to adapt the computation-intensive NeRF for high-resolution image synthesis, methods such as \cite{chan2022efficient,gu2021stylenerf,or2022stylesdf,xu20223d} first render posed 2D feature maps using a NeRF-based network and then generate high-resolution images in a two-stage pipeline.
In addition, efficient sampling techniques \cite{zhou2021cips,deng2022gram,xiang2022gram} are proposed to improve the efficiency for 3D-aware rendering.
Although high quality and view-consistent results can be synthesized from above methods, they often omit the semantics during the generation and lack the ability for part-level editing.

\textbf{Semantic-driven portrait images synthesis.}
Semantics is crucial for fined-grained and controllable image synthesis and editing.
Existing 2D GAN-based methods \cite{kim2021exploiting,kwon2021diagonal,shi2022semanticstylegan,xu2022transeditor,kafri2021stylefusion} have explored the utilization of semantics for disentangled 2D image editing.
To enable the 3D-aware semantic-driven portrait synthesis, recent efforts \cite{sun2022fenerf,chen2022sem2nerf,sun2022ide,zhang2022training,jiang2022nerffaceediting} incorporate the semantic masks 
to the generative learning process.
By simultaneous modeling the semantic masks and RGB images, methods such as IDE-3D \cite{sun2022ide} and NeRFFaceEditing \cite{jiang2022nerffaceediting} can achieve disentangled face editing by manipulating the semantic masks.
However, due to the holistic synthesis of the images and masks, their editing is still not strongly disentangled, i.e., when editing a certain local area, other parts may still be affected.

\textbf{Compositional portrait images synthesis.}
In contrast to the holistic image synthesis, compositional image synthesis can intrinsically ensure strong disentanglement between different image components or parts.
The composition could be conducted either at the image domain \cite{azadi2020compositional,sbai2021surprising,zhou2022} or by learning to generate and fuse implicit representations of individual components \cite{eslami2016attend,yang2017lr,burgess2019monet,greff2019multi,ehrhardt2020relate,arad2021compositional,niemeyer2021giraffe,xue2022giraffe,shi2022semanticstylegan}.
For compositional portrait synthesis, one representative work is SSGAN \cite{shi2022semanticstylegan} which learns to synthesize composed 2D face images and semantic masks from 2D part features generated by different 2D generators.
To enable 3D-aware compositional synthesis, CNeRF \cite{ma2023semantic} follows the idea of SSGAN but learns 3D generators to obtain 3D features and synthesize the posed 2D images and masks by volume aggregation and rendering.
Similar to CNeRF, our 3D-SSGAN also aims for 3D-aware compositional synthesis, while we leverage a depth-guided 2D-to-3D lifting module to obtain the 3D features.
3D-SSGAN can be easily trained by jointly optimizing the 2D generators and a semantic-aware volume renderer, and the results show strong part disentanglement while preserving 3D view consistency.

\section{Method}

Our 3D-SSGAN integrates the compositional synthesis design similar to SSGAN \cite{shi2022semanticstylegan} with the NeRF-based volume rendering \cite{xu20223d} for part-disentangled 3D-aware synthesis.
The overview of our training pipeline is shown in Fig.~\ref{pipeline}.
In the following, we first summarize the preliminaries on disentangled feature generation and NeRF-based rendering.
Then, we present the details of our framework.


\subsection{Preliminaries}
\label{pre}

\textbf{Disentangled feature generation.}
As proposed in SSGAN, 2D part features can be generated in a strongly disentangled manner.
First, a MLP is used to map $z$, which is randomly sampled from a standard normal distribution, into W space.
Then, W is further expanded into $W^{+}$ which contains latents for each semantic part.
Specifically, $W^{+}$ can be factorized as:
\begin{equation}
\label{eq:feature}
    W^{+}=W^{base}\times W^{1}_{s}\times W^{1}_{t}\times W^{2}_{s}\times W^{2}_{t} \cdots \times W^{K}_{s}\times W^{K}_{t},
\end{equation}%
where $W^{base}$ is a latent shared by all semantic parts to ensure the consistency in coarse structure;
$W^k_s$ and $W^k_t$ are corresponding to the shape and texture of $k$-th semantic part $p_k$, respectively; $K$ is the total number of semantic parts.
Independent 2D generators can then be used to generate disentangled part features from each $W^{k}=\{W^{base}, W^{k}_{s},W^{k}_{t}\}$.


\textbf{NeRF-based rendering.}
The key for 3D-aware synthesis is to use a neural or volume renderer to render images or 2D features from 3D features under given camera poses.
NeRF \cite{mildenhall2021nerf} is a popular neural renderer that can generate impressive novel view images.
To render a 3D volume, given a camera pose, a ray $\bm{r}(t)=\bm{o}+t\bm{d}$ with direction $\bm{d}$ is shooting from camera origin $\bm{o}$ and a set of points are sampled along the ray.
For each sampled point $X_i$, its radiance $c_i$ and density $\sigma_i$ are predicted using MLPs.
The final color of the pixel corresponding to the ray can be calculated as:
\begin{equation}
\label{eq:nerf_w}
    \hat{C}(\bm{r})=\sum_{i=1}^N w_{i}c_{i} {\rm\ ,\  where}\  w_{i}=T_{i}(1-{\rm exp}(-\sigma_{i}\delta_{i})),
\end{equation}
\begin{equation}
    T_{i}={\rm exp}\left(-\sum_{j=1}^{i-1}\sigma_{j}\delta_{j}\right).
\end{equation}%
Here, $\delta_i = t_{i+1}-t_i$ is the distance between adjacent sampled points, $w_i$ is the weight of each sampled 3D point and N is the total number of sampled points.
Inspired by NeRF, methods like VolumeGAN \cite{xu20223d} further synthesize 2D feature maps by performing NeRF-like volume rendering to 3D features generated by 3D generators: 
\begin{equation}
    f_{render}(\bm{r})=\sum_{i=1}^N w_{i}f_{i}^{3d}, 
\end{equation}%
where $w_i$ is the NeRF-based weights, $f_{i}^{3d}$ is the sampled 3D feature for point $X_i$ and $f_{render}$ is the rendered 2D feature.

\subsection{Feature Generation}
Different from existing methods \cite{sun2022ide,xu20223d,deng2022gram,jiang2022nerffaceediting,niemeyer2021giraffe,sun2022fenerf} which obtain 3D features by learning 3D generators, we lift 2D features generated by 2D generators. The lifted feature will be further optimized in the rendering and discriminative training process.

\begin{figure}[t]
    \centering
    \includegraphics[width=\columnwidth]{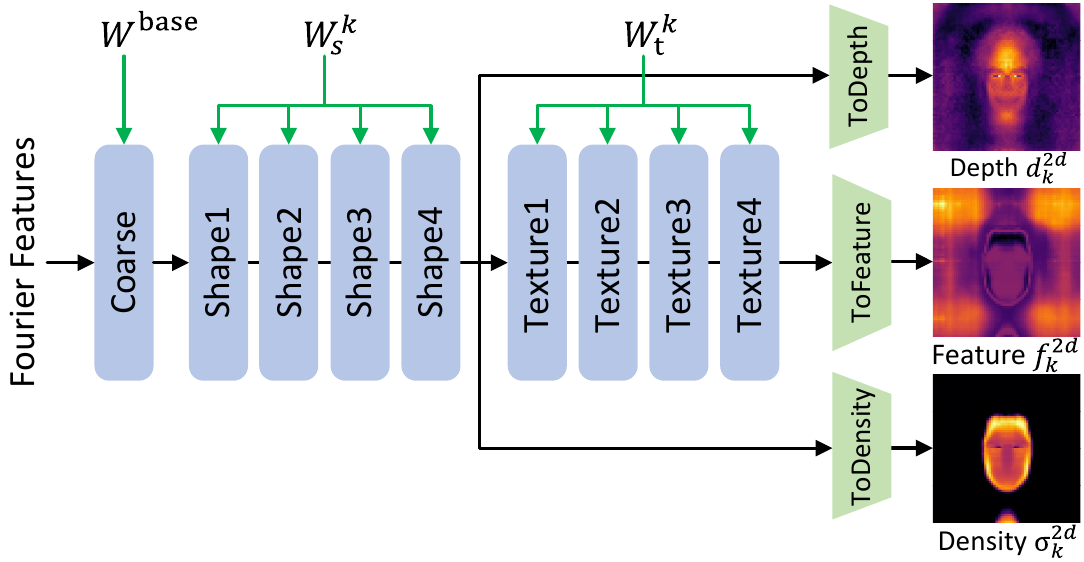}
    \caption{The architecture of our 2D generator. The Fourier features are 64-channel constant vectors, which work as position embedding. The blue blocks are $1\times 1$ convolution layers with 64 channels. 
    The green blocks are linear transformation layers.
    }
    \label{fig:local}
\end{figure}

\textbf{2D generator.}
\label{local}
First, we learn a 2D generator $G_k$ as shown in Fig.~\ref{fig:local} for each semantic part $p_k$.
Each $G_k$ takes the Fourier features, the generated latent $W^k$ and outputs a 2D feature map $f_k^{2d}$, 2D pseudo-depth map $d_k^{2d}$ and 2D density map $\sigma_k^{2d}$.
The overall architecture of $G_i$ is similar to the 2D generator in SSGAN \cite{shi2022semanticstylegan}, while we make three modifications to make it more suitable for the 3D-aware synthesis in later stages.
First, we use a one-layer MLP to replace the original two-layer MLP for learning the coarse structure, as we find the diversity of the results may be reduced, possibly due to the over constraints introduced by the two-layer MLP.
Secondly, for the depth map, to ease the learning and enhance the part disentanglement, we predict the relative depth w.r.t the part $p_{faceBase}$ for the facial parts normally above the face base.
For the part $p_{faceBase}$ and $p_{background}$, we directly predict their absolute depth values.
Lastly, in addition to the original ToFeature and ToDepth branch, we add a new branch ToDensity that generates a 2D density map $\sigma_k^{2d}$ for each part.
The 2D density map can be understood as the pseudo or soft semantic mask for the part and it can be further used to generate the 3D density volume for the following NeRF-based volume rendering.


\textbf{Depth-guided 2D-to-3D lifting module.}
\label{2to3}
To obtain the 3D features for volume rendering, we directly lift the generated 2D features and density maps to 3D with the guidance from the generated depth maps.
Considering a semantic part $p_k$, the lifting operations are defined as follows:
\begin{equation}
    f^{3d}_k(x,y,z) = \psi_k(x,y,z)f^{2d}_k(x,y),
\end{equation}
\begin{equation}
    \sigma^{3d}_k(x,y,z) = \psi_k(x,y,z)\sigma^{2d}_k(x,y),
\end{equation}
\begin{equation}
\label{eq:gauss}
    {\rm where\ }\psi_k(x,y,z) = {\rm exp}(-\alpha(\hat{d}^{2d}_k(x,y)-z)^2).
\end{equation}%
Here, $x,y \in \{0,1,2,...,63\}$ are the pixel position from the 2D map and $z\in \{0,1,2,...,31\}$ is the points uniformly sampled in the depth dimension.
Hence, the 3D feature volume $f^{3d}_k$ for each part $p_k$ is modeled as a $64 \times 64 \times 32$ grid.
$\psi_k$ is a depth-guided 2D-to-3D mapping function in the form of a general Gaussian function with $\alpha$ as a parameter that controls the width of the Gaussian ``bell''.
Given a 3D point $(x,y,z)$, $\psi_k(x,y,z)$ computes a weight based on the deviation between the point's $z$ value and its absolute depth $\hat{d}^{2d}_k(x,y)=d^{2d}_k(x,y)+d_{faceBase}$.
Then, the 3D feature and density values can be obtained by using $\psi_k$ to weight the corresponding 2D values.
Intuitively, our 2D-to-3D lifting can be understood as a \textit{soft unprojection} of the 2D features or densities to 3D space, while a closer distance between the $z$ value of the predicted depth, a higher weight is assigned.
Note that as the lifted 3D features will be further optimized in later stage, the specific mapping function should not be a critical factor for the final performance.
We verify this observation in the ablation study.


\subsection{Volume Rendering and Image Synthesis}
With the 3D feature volume obtained for each semantic part, we first fuse them into a global 3D feature volume and then render the 2D feature map similar to VolumeGAN \cite{xu20223d}.
In addition, the 2D semantic mask is generated by a novel 3D-aware semantic mask renderer.
Finally, the high-resolution image and mask are obtained from rendered 2D feature map and mask.

\textbf{NeRF-based 2D feature rendering.}
For a 3D point $(x,y,z)$, its 3D feature fused from different part features can be easily obtained by:
\begin{equation}
f^{3d}_{fused}(x,y,z) = \sum_{k=1}^K f_k^{3d}(x,y,z).
\end{equation}
The 3D density volume $\sigma^{3d}_{fused}$ can also be computed in a similar way.
Then, given a camera pose, a 2D feature map $f_{render}$ could be obtained by a feature renderer which is implemented following the pipeline introduced in Sec. \ref{pre}.

\textbf{3D-aware semantic mask renderer.}
To generate the semantic mask for a given camera pose, we independently render the mask $m_k$ for each semantic part so that the learned parts are more disentangled.
The process is similar to the feature rendering, while one main difference is, instead of sampling from the fused feature volume $f^{3d}_{fused}$, the 3D density volume $\sigma^{3d}_k$ for each part $p_k$ is sampled.
Specifically, the mask $m_k$ for a 2D pixel $(u,v)$ can be computed as:
\begin{equation}
\label{eq:softmax}
    m_k(u,v) = \frac{{\rm exp}(m_k(u,v))}{\sum_{j=1}^K {\rm exp}(m^{init}_j(u,v))},
\end{equation}
\begin{equation}
\label{eq:mask}
    m^{init}_k(u,v) = \sum_{i=1}^N w_i(u,v) \sigma_k^i(u,v),
\end{equation}
\begin{equation}
    \sigma_k^i(u,v) = \bm{SAMPLE}(\sigma_k^{3d},X_i(u,v)).
\end{equation}%
Here, $X_i(u,v)$ is the $i$-th 3D point sampled from the ray corresponding to pixel $(u,v)$, $\sigma_k^i(u,v)$ is the density value sampled (via interpolation) at $X_i(u,v)$, from the $k$-th 3D density volume $\sigma_k^{3d}$; 
$m^{init}_k(u,v)$ is the initial mask for part $p_k$, obtained by weighted averaging the density $\sigma_k^i(u,v)$ for each 3D point and $m_k(u,v)$ is computed by performing a softmax function to $m^{init}_k(u,v)$.
Following SSGAN \cite{shi2022semanticstylegan}, the final mask for $(u,v)$ is represented by a vector $\mathbf{m}(u,v)=\{m_1(u,v),m_2(u,v),\cdots,m_K(u,v)\}$. 
Note that by using the NeRF-based weights $w_i(u,v)$ defined in Eq. \ref{eq:nerf_w} to compute weighted average of the sampled density for each part, the 3D information learned from the fused 3D density volume $\sigma^{3d}_{fused}$ is considered, making our semantic mask renderer more 3D-aware.

\textbf{Final image and mask synthesis.}
Following SSGAN, we adopt a render net $R$ to generate the high-resolution image and mask.
The input is the rendered feature map $f_{render}$, and the output is a rendered high-res image and a mask residual $\Delta \mathbf{m}$ which can be used to get the final high-res mask $\mathbf{m_{hr}}=upsample(\mathbf{m})+\Delta \mathbf{m}$.


\subsection{Training and Implementation Details.}
To train our framework, we follow the adversarial training design from SSGAN \cite{shi2022semanticstylegan} and use a discriminator $D$ on the synthesized images and their corresponding masks. 
Besides the loss from SSGAN, a depth smoothness loss is considered to provide more constraints for the depth learning.
Specifically, our final loss is defined as:
\begin{equation}
\label{eq:loss}
    \mathcal{L}_{3D-SSGAN} = \mathcal{L}_{SSGAN} + \lambda_{ds}\mathcal{L}_{ds} {\rm\ ,\  where}\
\end{equation}%
\begin{equation}
    \mathcal{L}_{ds} = \frac{1}{A} \sum_{k} \sum_{x,y}\sum_{\delta_1,\delta_2}\Vert d_k^{2d}(x,y) - d_k^{2d}(x+\delta_1,y+\delta_2)\Vert^2.
\end{equation}
Here, $\mathcal{L}_{SSGAN}$ denotes the loss from SSGAN, which contains the original StyleGAN2 loss \cite{karras2020analyzing}, the mask loss and the regularization loss.
More details of $\mathcal{L}_{SSGAN}$ can be found in \cite{shi2022semanticstylegan}.
For $\mathcal{L}_{ds}$, the depth difference between neighboring pixels (specified by $\delta_1,\delta_2  \in \{-1,0,1\}$) in each part's depth map is summed up and averaged by a normalization term $A=8\cdot K\cdot H\cdot W$, where $8$ is the number of neighbors, $H$ and $W$ are the height and with of the depth map.
$\lambda_{depth}$ is a hyperparameter to control the strength of depth smoothness and is empirically set to 0.1 in our current implementation.

In our 2D generator, the ToFeature layer is the linear transformation layer which has the same architecture as in SSGAN and the output 2D feature $f^{2d}_k$ is $64\times 64 \times 256$.
The ToDensity and ToDepth layer are similar to the ToFeature, but with the output map as $64\times 64 \times 1$.
For training the volume renderer, the camera pose is sampled from a Gaussian distribution with mean of $\pi/2$, standard deviations of 0.3 and 0.155 for yaw and pitch, respectively.
The number of sampled points $N$ for each ray is 12 for training and 36 for testing to speed up the training, while trying to generate high-quality results for testing.
When training the 3D-aware semantic mask renderer, the NeRF-based weights for averaging the sampled density (Eq. \ref{eq:mask}) are not used for the first 2,000 iterations as the weights are not well-trained in the early training stage.

Our model is trained on the CelebAMask-HQ dataset \cite{lee2020maskgan} which contains 30K facial images labeled with 19 types of semantic segmentation masks. 
We use 13 types of the masks corresponding to portrait regions and feed the model with paired images and their segmentation masks as the real images for learning the discriminator.
The resolution of synthesized images and masks is 512$\times$512.
We implement our model in PyTorch and train it on a single NVIDIA A10 24G GPU with the batch size set to 4.
The current training time is about 109 hours and it could be substantially shortened when larger batch size and more GPUs are used.

\section{Results}

\begin{figure*}
  \centering
  \includegraphics[width=\textwidth]{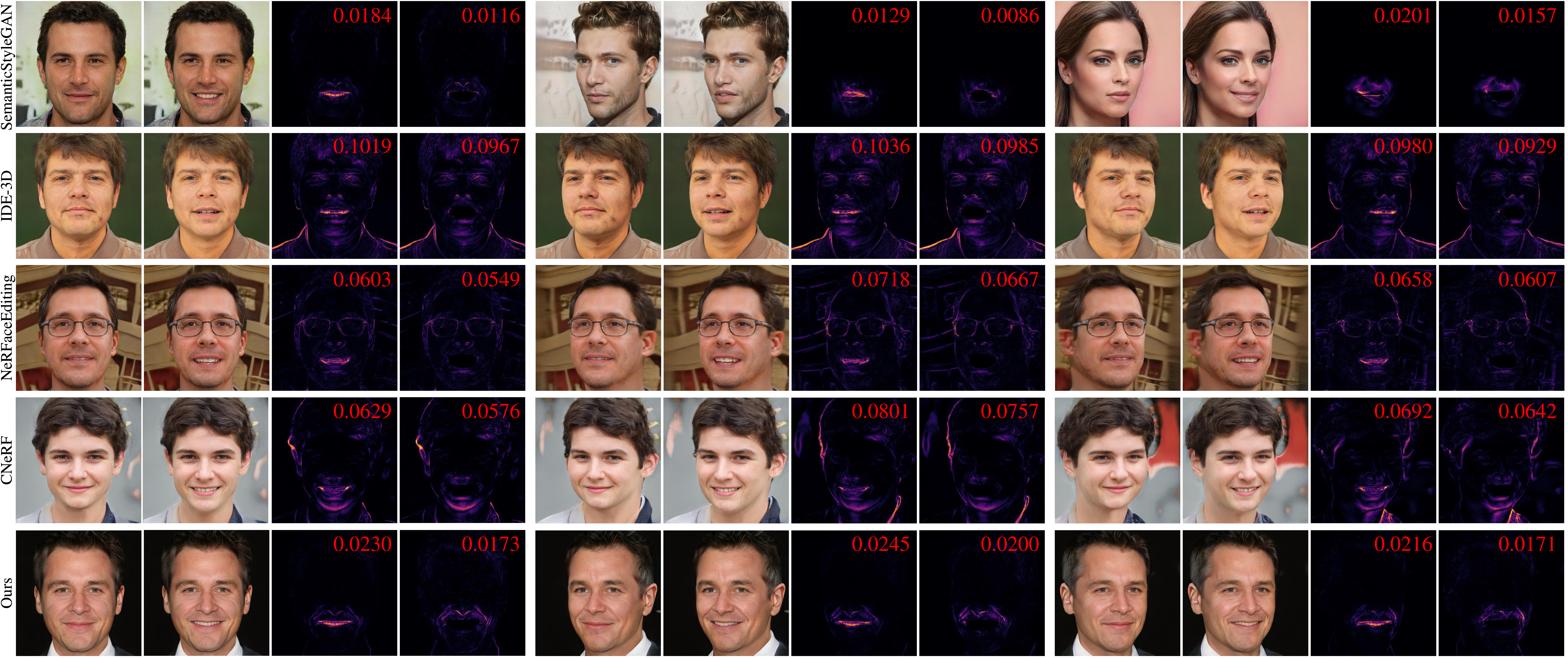}
  \caption{
  For each image group, part-level editing to the mouth region is applied to the left image generated by one portrait synthesis method and the result is shown on the right.
  The difference maps which encode the averaged difference of R, G, B channels are visualized, while the left map is the direct difference between two portrait images, and the right map is obtained by removing the mouth region using the corresponding part mask.
  The mean value of the whole difference map, i.e., $D_{mean}$ or $D_{mean}^{masked}$ is shown on the top right corner.
  It can be seen our method outperforms the 3D-aware synthesis methods (IDE-3D, NeRFFaceEditing, CNeRF) in disentangled part editing, while approaching similar performance to the 2D-only synthesis method SemanticStyleGAN.
  }
  \label{fig:diff}
\end{figure*}

\begin{figure*}[t]
    \centering
    \includegraphics[width=\textwidth]{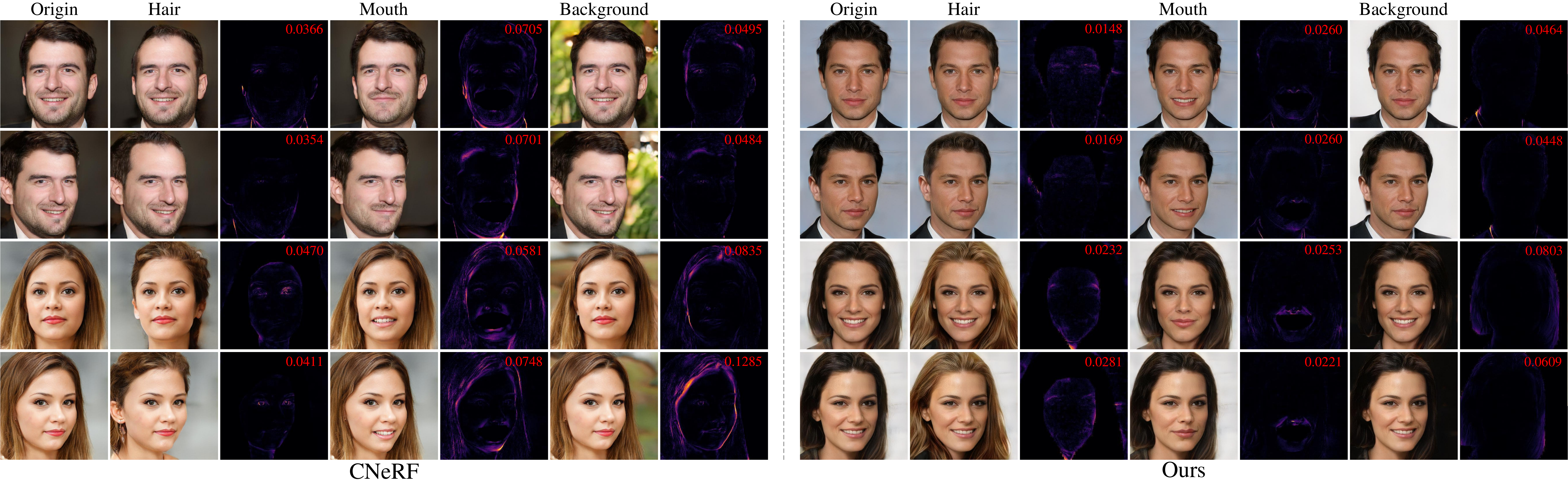}
    \caption{More comparisons with CNeRF in semantic-disentangled part-level editing. The difference maps, which can reveal the influence to other regions, are computed in the same way as in Fig.~\ref{fig:diff}. It can be seen our method has smaller influence to the non-edited regions and the superiority is consistant in different views.}
    \label{fig:cnerf}
\end{figure*}

In this section, we quantitatively and qualitatively evaluate our 3D-SSGAN, show its potential applications and perform ablation studies to evaluate our key modules. 

\subsection{Evaluation}

\textbf{Evaluation on 3D-aware semantic-disentanglement.}
The key advantage of our method is to achieve strong semantic-disentangled synthesis while preserving the 3D view consistency.
To evaluate its performance, we compare our method with SSGAN \cite{ma2023semantic}, IDE-3D \cite{sun2022ide}, NeRFaceEditing \cite{jiang2022nerffaceediting} and CNeRF \cite{ma2023semantic}, which are state-of-the-art methods for semantic-disentangled synthesis.
Among these methods, only CNeRF can intrinsically support strong semantic-disentangled and 3D-aware synthesis.
However, as CNeRF learns a 3D generator for each part, it is more memory and computation intensive than our method.
Fig.~\ref{fig:diff} shows qualitative and quantitative comparisons for the portrait synthesis and part-level editing.
For the quantitative comparison, we first compute the different maps, which encode the averaged difference of R, G, B channels between the original and edited images, to show the changes caused by the editing.
Then, the mean value $D_{mean}$ of the difference map is calculated as follows:
\begin{equation}
\resizebox{\columnwidth}{!}{
 $D_{mean}=\frac{1}{3\cdot W\cdot H} \sum_{c\in \left \{R, G, B  \right \} }^{} \sum_{i=1}^{W}\sum_{j=1}^{H}\left | p_{edited}(i,j,c)-p_{original} (i,j,c)  \right |$.
 }
\end{equation}%
Here, $p_{edited}(\cdot,\cdot,\cdot)$ and $p_{original}(\cdot,\cdot,\cdot)$ are the pixel values at the $(i,j)$ location in the $c$ channel of the edited and original images, respectively.
A smaller value of $D_{mean}$ indicates smaller changes on the whole images caused by the editing.
To further verify the changes for the non-edited regions, we also compute the difference map for the non-edited regions (i.e., a masked difference map, where the mask is the edited region) and denote its mean value as $D_{mean}^{masked}$.
For example, if the editing is conducted on the mouth region, the $D_{mean}^{masked}$ is computed using the difference values for the regions excluding the mouth.
Hence, $D_{mean}^{masked}$ can be a more suitable metric for evaluating the performance for the semantic-disentangled editing.
It can be observed our method achieves the best performance in disentangled part editing comparing to the 3D-aware methods IDE-3D, NeRFaceEditing, and CNeRF, while approaching similar performance to the 2D-only synthesis method SSGAN.

Furthermore, we conducted more quantitative and qualitative comparisons with CNeRF, which also supports strong semantic-disentangled and 3D-aware synthesis.
In Table \ref{tab:diff}, we show a relatively large-scale quantitative comparisons for the difference maps computed after editing different face parts. 
Specifically, we first generated 1,000 random portrait images. 
Then, for each of the interested face parts (i.e., mouth, eye, brow, nose and hair), we replaced the corresponding latent code with randomly generated latent code. 
Subsequently, the mean value of their respective difference maps was computed.
It can be found the differences caused by the editing are much smaller for our method than CNeRF for most cases, showing the superiority of our method in semantic-disentangled editing.
One exception is for the editing on the nose region, and the reason could be that the nose has a larger 3D depth variation within a small area, while a 3D generator might be more suitable for this case.
Fig.~\ref{fig:cnerf} shows some visual comparisons with CNeRF.
It can be seen that our method consistently surpasses CNeRF in semantic-disentangled part-level editing.
This should be due to the 3D generators used by CNeRF is harder to learn than our 2D generators, which produces the 2D features and obtaining the 3D features via lifting.

\begin{table}[t]
    \centering
    \resizebox{1\columnwidth}{!}{
    \renewcommand\tabcolsep{3.0pt}
    \begin{tabular}{l l c c c c c}
    \hline
     & Method & Mouth & Eye & Brow & Nose & Hair \\
       \hline
    \multirow{2}*{\makecell{$D_{mean}$}} 
     & CNeRF \cite{ma2023semantic} & 0.0818 & 0.0509&0.0452&\textbf{0.0583}&0.3111 \\
     &Ours& \textbf{0.0413}& \textbf{0.0340} & \textbf{0.0355}&0.0891&\textbf{0.2244}\\
      \hline
    \multirow{2}*{\makecell{$D_{mean}^{masked}$}}
    & CNeRF \cite{ma2023semantic} & 0.0658 & 0.0479&0.0420&\textbf{0.0552}&0.1665 \\
    &Ours& \textbf{0.0381}& \textbf{0.0334} & \textbf{0.0225}&0.0836&\textbf{0.0848} \\     
    \hline
    \end{tabular}
    }
    \caption{
    Quantitative comparison with CNeRF on semantic-disentanglement. Mean value of the difference map and masked difference map, $D_{mean}$ and $D_{mean}^{masked}$, are computed and compared for each interested region.
    }
    \label{tab:diff}
\end{table}

\begin{figure}[t]
    \centering
    \includegraphics[width=.48\textwidth]{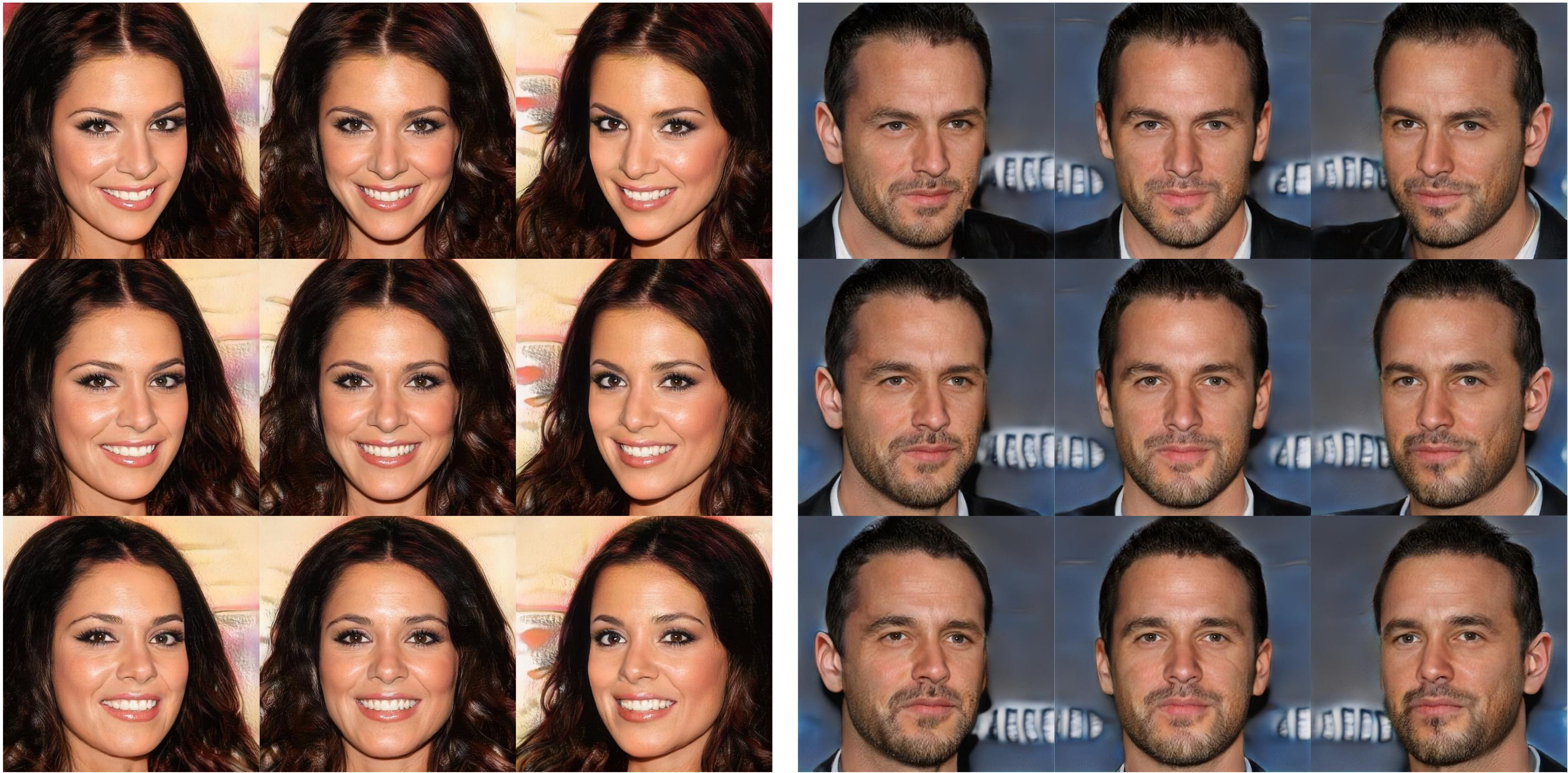}
    \caption{High quality 3D view-consistent images synthesized by specifying different camera poses for our method. 
    }
    \label{pose}
\end{figure}

\textbf{Evaluation on synthesis quality.}
We conduct qualitative and quantitative evaluation on the quality of the synthesized portrait images.
Fig.~\ref{fig:teaser} shows the qualitative results of 3D-SSGAN for 3D-aware part-level synthesis.
It can be seen our method can achieve semantic-disentangled editing while preserving view consistency.
Moreover, compared with a 2D GAN method such as SemanticStyleGAN, our model could generate high-quality 3D-consistent images from different view points, as shown in Fig.~\ref{pose}, which verify the effectiveness of our 3D feature representation generated via 2D-to-3D lifting.
To quantitatively evaluate the quality of synthesized images, we compute the Frechet Inception Distance (FID) \cite{heusel2017gans} between 30K CelebAMask-HQ images and 50K images generated by our approach. 
The camera pose corresponding to each of our images is sampled from a Gaussian distribution.
Table \ref{tab:fid} shows a preliminary comparison on the quality of images synthesized by relevant methods.
Note all compared models are trained CelebAMask-HQ.
Other models such as CNeRF\cite{ma2023semantic} are not compared since they do not report the FID results on CelebAMask-HQ.
It can be observed our FID is plausible and the performance is close to FENeRF \cite{sun2022fenerf}, which is also a representative 3D-aware and semantic-driven synthesis method.
Meanwhile, although the compared methods produce better FIDs, they can not achieve the strong semantic-disentangled and 3D-aware synthesis as our 3D-SSGAN (see Fig.~\ref{fig:diff}, \ref{fig:cnerf} and Table \ref{tab:diff}).

\begin{table}[t]
    \centering
    \resizebox{1\columnwidth}{!}{
    \begin{tabular}{l c c c}
    \hline
    Method & Semantic-disentanglement & 3D-aware & FID$\downarrow$ \\
    \hline
    Ours&Strong & \ding{52} & 14.5 \\
    SSGAN \cite{shi2022semanticstylegan} & Strong & \ding{56} & 5.7 \\
    IDE-3D \cite{sun2022ide} & Weak & \ding{52} & 4.9 \\
    FENeRF \cite{sun2022fenerf} & Weak & \ding{52} & 12.1 \\
    \hline
    \end{tabular}
    }
    \caption{
    Preliminary quantitative comparison on the quality of synthesized images.
    It can be seen our method achieves a plausible FID while maintaining both strong semantic-disentanglement and 3D-aware property.
    }
    \label{tab:fid}
\end{table}

\begin{figure}[t]
    \centering
    \includegraphics[width=\columnwidth]{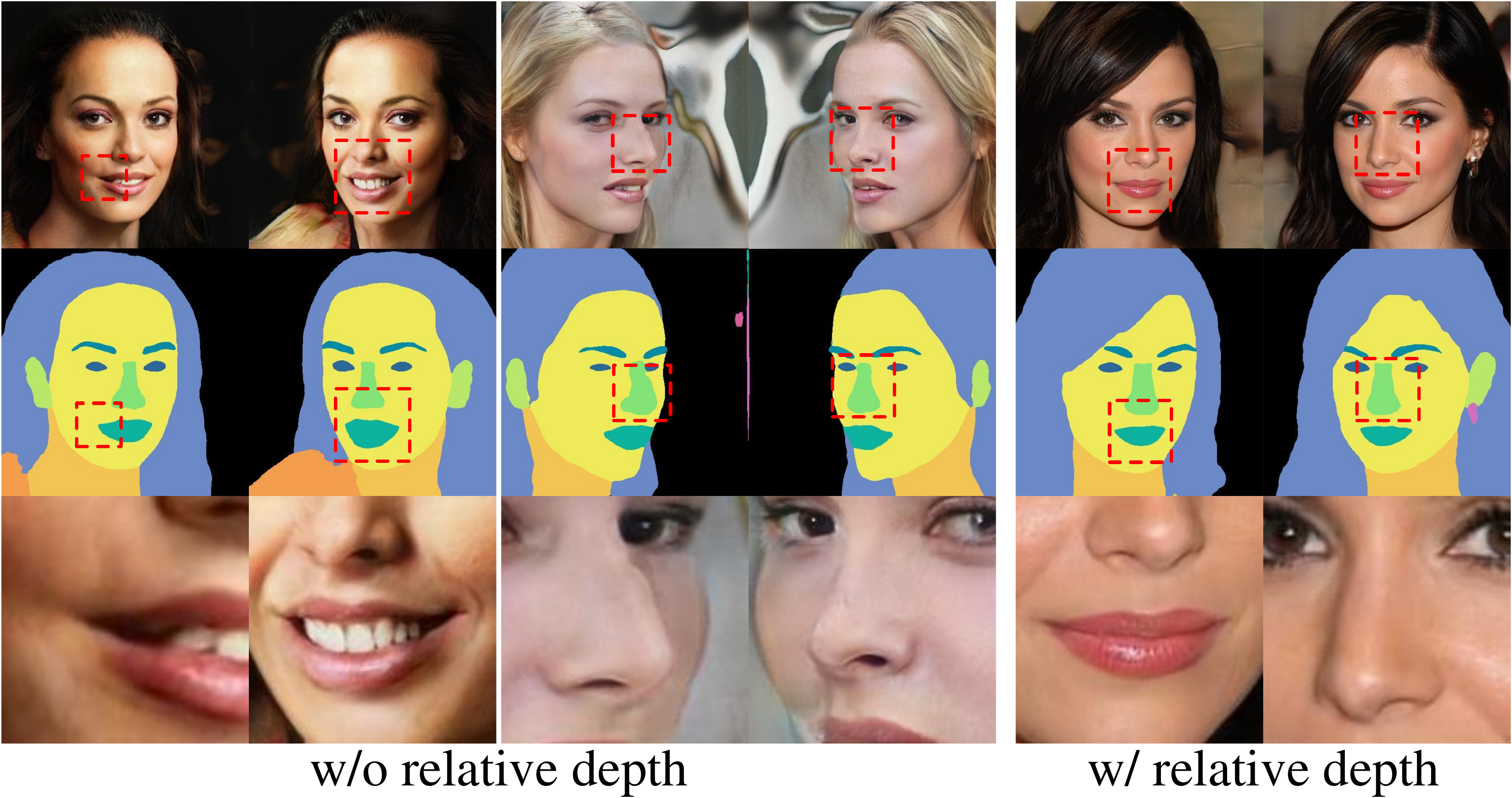}
    \caption{Effectiveness of learning relative depth in 2D generators. When using relative depth, better 3D relationships  are learned between parts, especially for the nose and mouth region.}
    \label{relative}
\end{figure}

\begin{table}[t]
    \centering
    \begin{tabular}{l c c c}
    \hline
    Method &  FID$\downarrow$ \\
    \hline
    Full model& 14.5 \\
    w/o relative depth & 20.01\\
    w/o NeRF-based weights & 47.48 \\
    \hline
    \end{tabular}
    \caption{Quantitative comparison of ablation studies.}
    \label{tab:ablation}
\end{table}

\subsection{Ablation Studies}
\label{ablation}

\textbf{Learning relative depth in 2D generators.}
To facilitate the 3D-aware learning, we propose to learn the depth relative to the face base $p_{faceBase}$ instead of the absolute depth as the SS-GAN \cite{shi2022semanticstylegan} does.
By learning the relative depth, the 3D information, as well as the relationship between the semantic parts can be better learned.
From the comparison in Fig.~\ref{relative}, it shows for the nose and mouth region which contains moderately large depth variations, if the depth is not learned relatively, the synthesized images and masks may both be inconsistent across the views.
We also conducted quantitative comparisons between the full model and the model that does not use the relative depth in 2D generators.
The FID values between the 30K CelebAMask-HQ images and 50K images generated by each model are computed and the results are shown in Table \ref{tab:ablation}.
The second row in the table shows the FID considerably increases comparing to the full model, which means the quality or realism of the results drops if the relative depth is not used during the learning process.

\begin{figure*}[t]
    \centering
    \includegraphics[width=\textwidth]{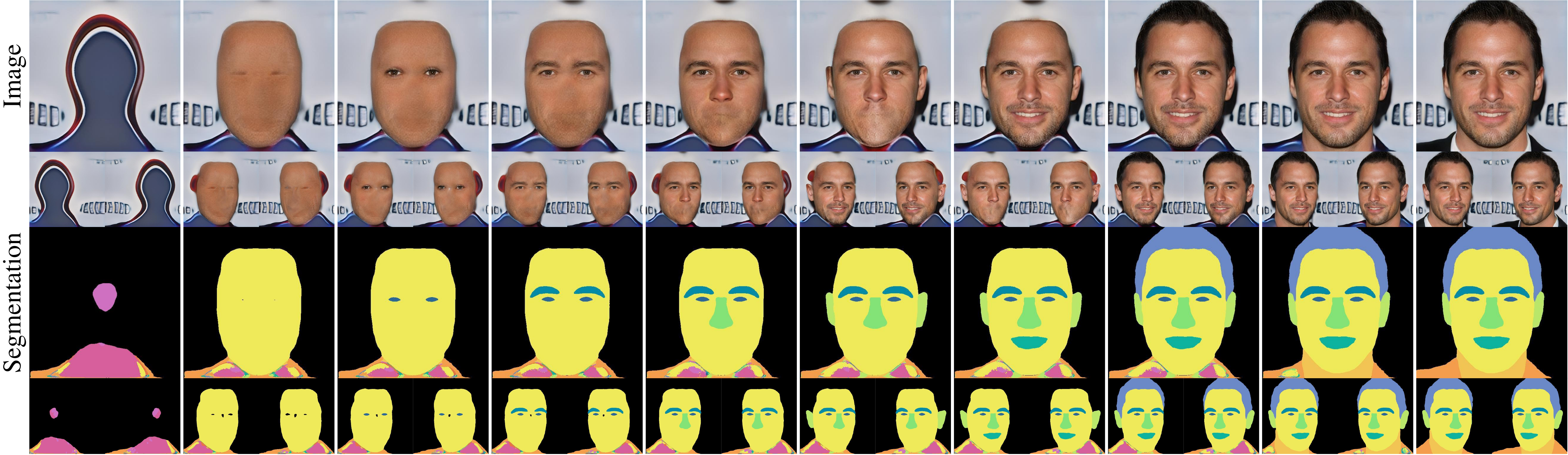}
    \caption{
    Progressive 3D-aware semantic-disentangled face generation. Our method is able to generate the semantic parts progressively, while 3D view-consistent images and semantic masks can be obtained at each step.
    }
    \label{fig:comp}
\end{figure*}

\begin{figure}[t]
    \centering
    \includegraphics[width=\columnwidth]{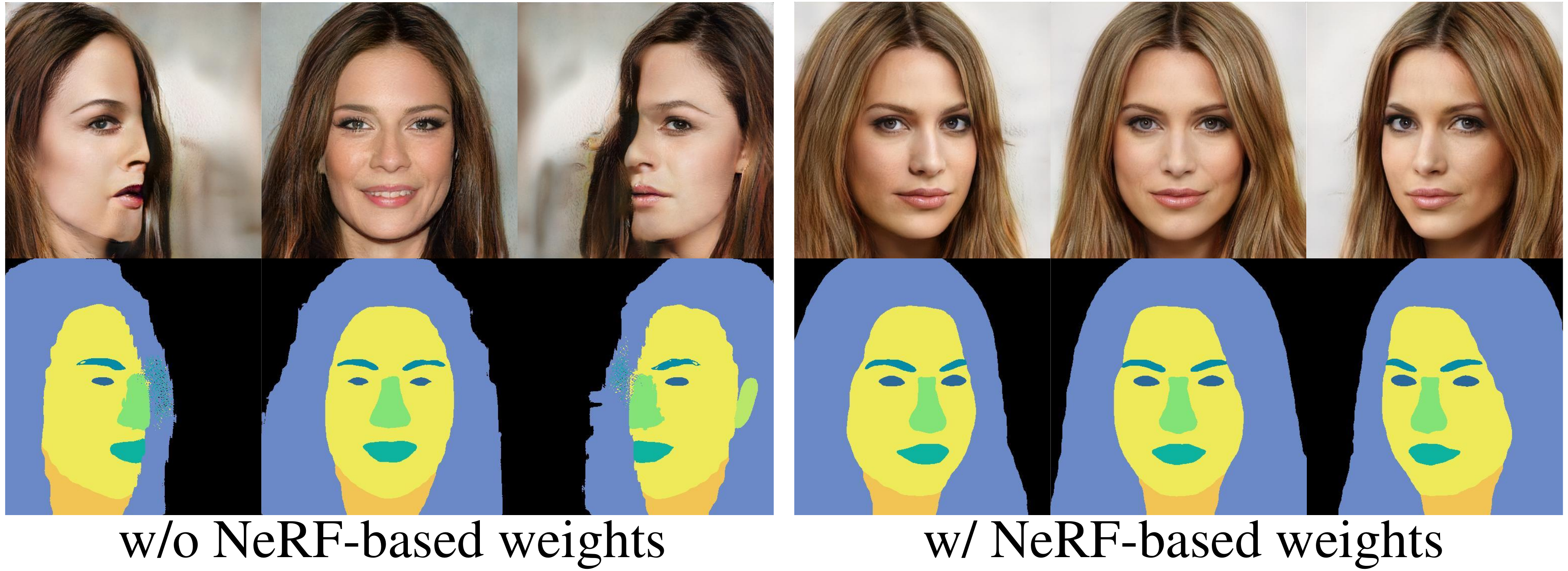}
    \caption{
    Effectiveness of using NeRF-based weights in the semantic-aware mask renderer. Without using the weights, the masks rendered from non-frontal views are implausible, as no 3D information is considered.
    }
    \label{occlusion}
\end{figure}



\textbf{3D-aware semantic mask renderer.}
To make the semantic mask rendering 3D-aware, we utilize the NeRF-based weights when combining the sampled density for each part.
Fig.~\ref{occlusion} shows the ablation study on the effectiveness of using NeRF-based weights during the mask rendering.
The baseline results are obtained by setting the weights $w_i$ to 1 in Eq. \ref{eq:mask}, meaning the density values are uniformly weighted instead of using the NeRF-based weights.
It can be seen without using the NeRF-based weights, only the results from the frontal view are plausible.
Also, we compare the FID values between the 30K CelebAMask-HQ images and 50K images generated by the full model and the model which does not use the NeRF-based weights for the smentic mask rendering.
In the last row of Table \ref{tab:ablation}, the FID significantly increases comparing to other models if the NeRF-based weights is not used.
This results quantitatively verify the importance of NeRF-based weights for generating realism 3D-aware results.

\textbf{2D-to-3D lifting strategy.}
The depth-guided 2D-to-3D lifting is our key idea for efficiently obtaining 3D features from 2D generators.
To achieve the lifting, we define the parametric Gaussian function $\phi_k$ (Eq. \ref{eq:gauss}) to \textit{soft-unproject} the 2D features and densities to 3D.
To verify the influence of different mapping functions, we train different models with $\alpha=1 \ or \ 2$, as well as other two functions which are symmetric inverse-proportional, and compare the FIDs of synthesized images.
The results produce consistent FID values which are around 14, showing the network can adapt to different lifting strategies, without being affected by specific mapping functions.

\subsection{Application}

\textbf{Progressive 3D-aware semantic-disentangled face generation.}
As our method is strongly semantic-disentangled, i.e., the semantic parts are generated by independent latent codes and 2D generators, we can enable progressive face generation by incrementally generating more parts.
Furthermore, 3D-aware images and corresponding semantic masks can also be generated along the process.
Fig.~\ref{fig:comp} shows an example of the progressive generation.
More applications such as controllable and 3D-aware part-level composition and editing can be enabled in a similar manner.

\begin{figure}[t]
    \centering
    \includegraphics[width=.48\textwidth]{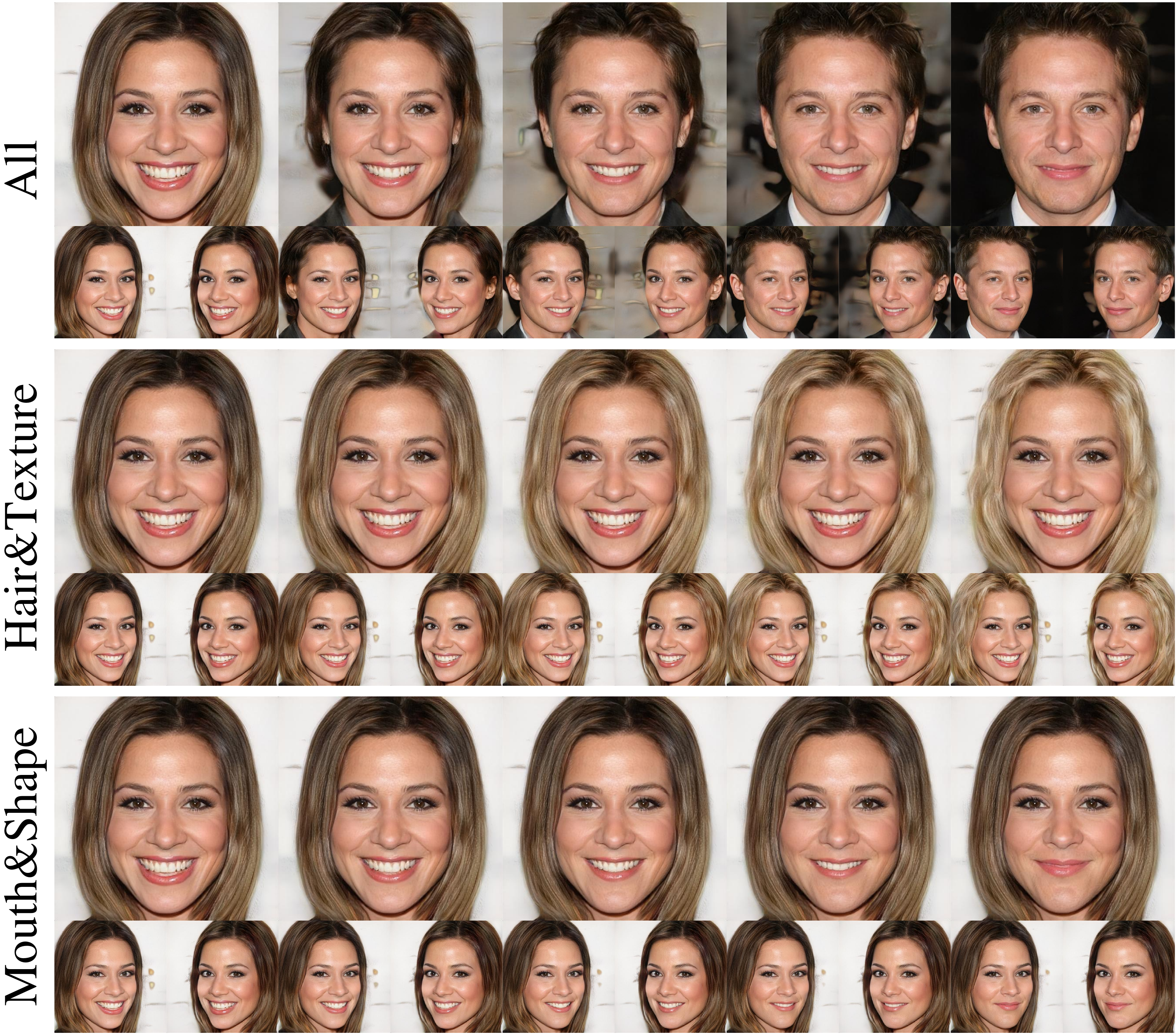}
    \caption{3D-aware part-level style interpolation. By interpolating the latent codes of specifying parts, smooth face transitions could be obtained, while the faces generated in each step are 3D view-consistent.}
    \label{fig:interpolate}
\end{figure}


\textbf{3D-aware part-level style interpolation.}
Similar to SemanticStyleGAN \cite{shi2022semanticstylegan}, our semantic-disentangled synthesis approach also supports face generation using the interpolated part features and enables the application part-level style interpolation.
Moreover, our faces generated at each step are 3D view-consistent.
Fig.~\ref{fig:interpolate} shows the results of faces generated by performing the latent interpolation. 
Given two randomly sampled images in the first row, our 3D-SSGAN can generate smoothly transited 3D-aware faces using the interpolated features for each part.
Besides, we can interpolate on a specific semantic area by changing the corresponding latent code (e.g., hair or mouth) while fixing irrelevant parts. 
Moreover, as the shape and texture of different local parts are disentangled (cf. Eq. \ref{eq:feature}), we can also choose which kind of features to interpolate.
The results in last two rows verify that our method has learned a smooth and disentangled latent space for controllable 3D-aware semantic editing. 

\section{Conclusion}


In this paper, we propose 3D-SSGAN, a novel framework for 3D-aware compositional portrait synthesis.
Our novel depth-guided 2D-to-3D lifting module simply extends the semantic-disentangled learning to be 3D-aware by a lifting operation.
With the lifted and fused features, the volume renderer is optimized to synthesize 3D-aware face images along with their semantic masks.
Moreover, the NeRF-based weights are utilized to during the part-level mask generation, making the generated faces more 3D-aware.
The 3D-aware semantic mask renderer also effectively incorporates the 3D information into the part-level mask generation.
Although our 3D-SSGAN demonstrates superior performance in strongly semantic-disentangled 3D-aware synthesis, there are still improvement room for the future work.
Firstly, more evaluations on other datasets such as FFHQ \cite{karras2019style} are needed to further verify the generalizability of our method.
Secondly, quality and diversity of the results may be further improved by adjusting the architecture of 2D generators.
Last but not least, how to further improve the memory and computation efficiency of our compositional-based network will be an interesting future direction.


\section*{Acknowledgments}
This work is supported in part by National Natural Science Foundation of China (62202199).

\bibliographystyle{cag-num-names}
\bibliography{refs}

\begin{thebibliography}{45}
\providecommand{\natexlab}[1]{#1}
\providecommand{\url}[1]{\texttt{#1}}
\providecommand{\href}[2]{#2}
\providecommand{\path}[1]{#1}
\providecommand{\eprint}[1]{\href{http://arxiv.org/abs/#1}{\path{#1}}}
\providecommand{\DOIprefix}{doi:}
\providecommand{\ArXivprefix}{arXiv:}
\providecommand{\URLprefix}{URL: }
\providecommand{\Pubmedprefix}{pmid:}
\providecommand{\doi}[1]{\href{http://dx.doi.org/#1}{\path{#1}}}
\providecommand{\Pubmed}[1]{\href{pmid:#1}{\path{#1}}}
\providecommand{\BIBand}{and}
\providecommand{\bibinfo}[2]{#2}
\ifx\xfnm\undefined \def\xfnm[#1]{\unskip,\space#1}\fi
\bibitem[{Karras et~al.(2019)Karras, Laine and Aila}]{karras2019style}
\bibinfo{author}{Karras\xfnm[ T]}, \bibinfo{author}{Laine\xfnm[ S]}, \bibinfo{author}{Aila\xfnm[ T]}.
\newblock \bibinfo{title}{A style-based generator architecture for generative adversarial networks}.
\newblock In: \bibinfo{booktitle}{Proceedings of the IEEE/CVF conference on computer vision and pattern recognition}. \bibinfo{year}{2019}, p. \bibinfo{pages}{4401--4410}.
\bibitem[{Karras et~al.(2020)Karras, Laine, Aittala, Hellsten, Lehtinen and Aila}]{karras2020analyzing}
\bibinfo{author}{Karras\xfnm[ T]}, \bibinfo{author}{Laine\xfnm[ S]}, \bibinfo{author}{Aittala\xfnm[ M]}, \bibinfo{author}{Hellsten\xfnm[ J]}, \bibinfo{author}{Lehtinen\xfnm[ J]}, \bibinfo{author}{Aila\xfnm[ T]}.
\newblock \bibinfo{title}{Analyzing and improving the image quality of stylegan}.
\newblock In: \bibinfo{booktitle}{Proceedings of the IEEE/CVF conference on computer vision and pattern recognition}. \bibinfo{year}{2020}, p. \bibinfo{pages}{8110--8119}.
\bibitem[{Karras et~al.(2021)Karras, Aittala, Laine, H{\"a}rk{\"o}nen, Hellsten, Lehtinen et~al.}]{karras2021alias}
\bibinfo{author}{Karras\xfnm[ T]}, \bibinfo{author}{Aittala\xfnm[ M]}, \bibinfo{author}{Laine\xfnm[ S]}, \bibinfo{author}{H{\"a}rk{\"o}nen\xfnm[ E]}, \bibinfo{author}{Hellsten\xfnm[ J]}, \bibinfo{author}{Lehtinen\xfnm[ J]}, et~al.
\newblock \bibinfo{title}{Alias-free generative adversarial networks}.
\newblock \bibinfo{journal}{Advances in Neural Information Processing Systems} \bibinfo{year}{2021};\bibinfo{volume}{34}:\bibinfo{pages}{852--863}.
\bibitem[{Bermano et~al.(2022)Bermano, Gal, Alaluf, Mokady, Nitzan, Tov et~al.}]{bermano2022}
\bibinfo{author}{Bermano\xfnm[ A]}, \bibinfo{author}{Gal\xfnm[ R]}, \bibinfo{author}{Alaluf\xfnm[ Y]}, \bibinfo{author}{Mokady\xfnm[ R]}, \bibinfo{author}{Nitzan\xfnm[ Y]}, \bibinfo{author}{Tov\xfnm[ O]}, et~al.
\newblock \bibinfo{title}{State-of-the-art in the architecture, methods and applications of stylegan}.
\newblock \bibinfo{journal}{Computer Graphics Forum} \bibinfo{year}{2022};\bibinfo{volume}{41}(\bibinfo{number}{2}):\bibinfo{pages}{591--611}.
\bibitem[{Shi et~al.(2022)Shi, Yang, Wan and Shen}]{shi2022semanticstylegan}
\bibinfo{author}{Shi\xfnm[ Y]}, \bibinfo{author}{Yang\xfnm[ X]}, \bibinfo{author}{Wan\xfnm[ Y]}, \bibinfo{author}{Shen\xfnm[ X]}.
\newblock \bibinfo{title}{Semanticstylegan: Learning compositional generative priors for controllable image synthesis and editing}.
\newblock In: \bibinfo{booktitle}{Proceedings of the IEEE/CVF Conference on Computer Vision and Pattern Recognition}. \bibinfo{year}{2022}, p. \bibinfo{pages}{11254--11264}.
\bibitem[{Gu et~al.(2021)Gu, Liu, Wang and Theobalt}]{gu2021stylenerf}
\bibinfo{author}{Gu\xfnm[ J]}, \bibinfo{author}{Liu\xfnm[ L]}, \bibinfo{author}{Wang\xfnm[ P]}, \bibinfo{author}{Theobalt\xfnm[ C]}.
\newblock \bibinfo{title}{Stylenerf: A style-based 3d-aware generator for high-resolution image synthesis}.
\newblock \bibinfo{journal}{arXiv preprint arXiv:211008985} \bibinfo{year}{2021};.
\bibitem[{Xu et~al.(2022{\natexlab{a}})Xu, Peng, Yang, Shen and Zhou}]{xu20223d}
\bibinfo{author}{Xu\xfnm[ Y]}, \bibinfo{author}{Peng\xfnm[ S]}, \bibinfo{author}{Yang\xfnm[ C]}, \bibinfo{author}{Shen\xfnm[ Y]}, \bibinfo{author}{Zhou\xfnm[ B]}.
\newblock \bibinfo{title}{3d-aware image synthesis via learning structural and textural representations}.
\newblock In: \bibinfo{booktitle}{Proceedings of the IEEE/CVF Conference on Computer Vision and Pattern Recognition}. \bibinfo{year}{2022}{\natexlab{a}}, p. \bibinfo{pages}{18430--18439}.
\bibitem[{Chan et~al.(2022)Chan, Lin, Chan, Nagano, Pan, De~Mello et~al.}]{chan2022efficient}
\bibinfo{author}{Chan\xfnm[ ER]}, \bibinfo{author}{Lin\xfnm[ CZ]}, \bibinfo{author}{Chan\xfnm[ MA]}, \bibinfo{author}{Nagano\xfnm[ K]}, \bibinfo{author}{Pan\xfnm[ B]}, \bibinfo{author}{De~Mello\xfnm[ S]}, et~al.
\newblock \bibinfo{title}{Efficient geometry-aware 3d generative adversarial networks}.
\newblock In: \bibinfo{booktitle}{Proceedings of the IEEE/CVF Conference on Computer Vision and Pattern Recognition}. \bibinfo{year}{2022}, p. \bibinfo{pages}{16123--16133}.
\bibitem[{Deng et~al.(2022)Deng, Yang, Xiang and Tong}]{deng2022gram}
\bibinfo{author}{Deng\xfnm[ Y]}, \bibinfo{author}{Yang\xfnm[ J]}, \bibinfo{author}{Xiang\xfnm[ J]}, \bibinfo{author}{Tong\xfnm[ X]}.
\newblock \bibinfo{title}{Gram: Generative radiance manifolds for 3d-aware image generation}.
\newblock In: \bibinfo{booktitle}{Proceedings of the IEEE/CVF Conference on Computer Vision and Pattern Recognition}. \bibinfo{year}{2022}, p. \bibinfo{pages}{10673--10683}.
\bibitem[{Or-El et~al.(2022)Or-El, Luo, Shan, Shechtman, Park and Kemelmacher-Shlizerman}]{or2022stylesdf}
\bibinfo{author}{Or-El\xfnm[ R]}, \bibinfo{author}{Luo\xfnm[ X]}, \bibinfo{author}{Shan\xfnm[ M]}, \bibinfo{author}{Shechtman\xfnm[ E]}, \bibinfo{author}{Park\xfnm[ JJ]}, \bibinfo{author}{Kemelmacher-Shlizerman\xfnm[ I]}.
\newblock \bibinfo{title}{Stylesdf: High-resolution 3d-consistent image and geometry generation}.
\newblock In: \bibinfo{booktitle}{Proceedings of the IEEE/CVF Conference on Computer Vision and Pattern Recognition}. \bibinfo{year}{2022}, p. \bibinfo{pages}{13503--13513}.
\bibitem[{Goodfellow et~al.(2014)Goodfellow, Pouget-Abadie, Mirza, Xu, Warde-Farley, Ozair et~al.}]{goodfellow2014generative}
\bibinfo{author}{Goodfellow\xfnm[ IJ]}, \bibinfo{author}{Pouget-Abadie\xfnm[ J]}, \bibinfo{author}{Mirza\xfnm[ M]}, \bibinfo{author}{Xu\xfnm[ B]}, \bibinfo{author}{Warde-Farley\xfnm[ D]}, \bibinfo{author}{Ozair\xfnm[ S]}, et~al.
\newblock \bibinfo{title}{Generative adversarial nets}.
\newblock \bibinfo{journal}{stat} \bibinfo{year}{2014};\bibinfo{volume}{1050}:\bibinfo{pages}{10}.
\bibitem[{Mildenhall et~al.(2021)Mildenhall, Srinivasan, Tancik, Barron, Ramamoorthi and Ng}]{mildenhall2021nerf}
\bibinfo{author}{Mildenhall\xfnm[ B]}, \bibinfo{author}{Srinivasan\xfnm[ PP]}, \bibinfo{author}{Tancik\xfnm[ M]}, \bibinfo{author}{Barron\xfnm[ JT]}, \bibinfo{author}{Ramamoorthi\xfnm[ R]}, \bibinfo{author}{Ng\xfnm[ R]}.
\newblock \bibinfo{title}{Nerf: Representing scenes as neural radiance fields for view synthesis}.
\newblock \bibinfo{journal}{Communications of the ACM} \bibinfo{year}{2021};\bibinfo{volume}{65}(\bibinfo{number}{1}):\bibinfo{pages}{99--106}.
\bibitem[{Sun et~al.(2022{\natexlab{a}})Sun, Wang, Shi, Wang, Wang and Liu}]{sun2022ide}
\bibinfo{author}{Sun\xfnm[ J]}, \bibinfo{author}{Wang\xfnm[ X]}, \bibinfo{author}{Shi\xfnm[ Y]}, \bibinfo{author}{Wang\xfnm[ L]}, \bibinfo{author}{Wang\xfnm[ J]}, \bibinfo{author}{Liu\xfnm[ Y]}.
\newblock \bibinfo{title}{Ide-3d: Interactive disentangled editing for high-resolution 3d-aware portrait synthesis}.
\newblock \bibinfo{journal}{arXiv preprint arXiv:220515517} \bibinfo{year}{2022}{\natexlab{a}};.
\bibitem[{Jiang et~al.(2022)Jiang, Chen, Liu, Fu and Gao}]{jiang2022nerffaceediting}
\bibinfo{author}{Jiang\xfnm[ K]}, \bibinfo{author}{Chen\xfnm[ SY]}, \bibinfo{author}{Liu\xfnm[ FL]}, \bibinfo{author}{Fu\xfnm[ H]}, \bibinfo{author}{Gao\xfnm[ L]}.
\newblock \bibinfo{title}{Nerffaceediting: Disentangled face editing in neural radiance fields}.
\newblock In: \bibinfo{booktitle}{SIGGRAPH Asia 2022 Conference Papers}. \bibinfo{year}{2022}, p. \bibinfo{pages}{1--9}.
\bibitem[{Ma et~al.(2023)Ma, Li, He, Dong and Tan}]{ma2023semantic}
\bibinfo{author}{Ma\xfnm[ T]}, \bibinfo{author}{Li\xfnm[ B]}, \bibinfo{author}{He\xfnm[ Q]}, \bibinfo{author}{Dong\xfnm[ J]}, \bibinfo{author}{Tan\xfnm[ T]}.
\newblock \bibinfo{title}{Semantic 3d-aware portrait synthesis and manipulation based on compositional neural radiance field}.
\newblock In: \bibinfo{booktitle}{Proceedings of the Thirty-Seventh AAAI Conference on Artificial Intelligence (AAAI)}. \bibinfo{year}{2023},.
\bibitem[{Tewari et~al.(2022)Tewari, Thies, Mildenhall, Srinivasan, Tretschk, Yifan et~al.}]{tewari2022}
\bibinfo{author}{Tewari\xfnm[ A]}, \bibinfo{author}{Thies\xfnm[ J]}, \bibinfo{author}{Mildenhall\xfnm[ B]}, \bibinfo{author}{Srinivasan\xfnm[ P]}, \bibinfo{author}{Tretschk\xfnm[ E]}, \bibinfo{author}{Yifan\xfnm[ W]}, et~al.
\newblock \bibinfo{title}{Advances in neural rendering}.
\newblock \bibinfo{journal}{Computer Graphics Forum} \bibinfo{year}{2022};\bibinfo{volume}{41}(\bibinfo{number}{2}):\bibinfo{pages}{703--735}.
\bibitem[{Xia and Xue(2022)}]{xia2022survey}
\bibinfo{author}{Xia\xfnm[ W]}, \bibinfo{author}{Xue\xfnm[ JH]}.
\newblock \bibinfo{title}{A survey on 3d-aware image synthesis}.
\newblock \bibinfo{year}{2022}.
\newblock \href{http://arxiv.org/abs/2210.14267}{\tt arXiv:2210.14267}.
\bibitem[{Zhu et~al.(2018)Zhu, Zhang, Zhang, Wu, Torralba, Tenenbaum et~al.}]{zhu2018visual}
\bibinfo{author}{Zhu\xfnm[ JY]}, \bibinfo{author}{Zhang\xfnm[ Z]}, \bibinfo{author}{Zhang\xfnm[ C]}, \bibinfo{author}{Wu\xfnm[ J]}, \bibinfo{author}{Torralba\xfnm[ A]}, \bibinfo{author}{Tenenbaum\xfnm[ J]}, et~al.
\newblock \bibinfo{title}{Visual object networks: Image generation with disentangled 3d representations}.
\newblock \bibinfo{journal}{Advances in neural information processing systems} \bibinfo{year}{2018};\bibinfo{volume}{31}.
\bibitem[{Henzler et~al.(2019)Henzler, Mitra and Ritschel}]{henzler2019escaping}
\bibinfo{author}{Henzler\xfnm[ P]}, \bibinfo{author}{Mitra\xfnm[ NJ]}, \bibinfo{author}{Ritschel\xfnm[ T]}.
\newblock \bibinfo{title}{Escaping plato's cave: 3d shape from adversarial rendering}.
\newblock In: \bibinfo{booktitle}{Proceedings of the IEEE/CVF International Conference on Computer Vision}. \bibinfo{year}{2019}, p. \bibinfo{pages}{9984--9993}.
\bibitem[{Nguyen-Phuoc et~al.(2019)Nguyen-Phuoc, Li, Theis, Richardt and Yang}]{nguyen2019hologan}
\bibinfo{author}{Nguyen-Phuoc\xfnm[ T]}, \bibinfo{author}{Li\xfnm[ C]}, \bibinfo{author}{Theis\xfnm[ L]}, \bibinfo{author}{Richardt\xfnm[ C]}, \bibinfo{author}{Yang\xfnm[ YL]}.
\newblock \bibinfo{title}{Hologan: Unsupervised learning of 3d representations from natural images}.
\newblock In: \bibinfo{booktitle}{Proceedings of the IEEE/CVF International Conference on Computer Vision}. \bibinfo{year}{2019}, p. \bibinfo{pages}{7588--7597}.
\bibitem[{Nguyen-Phuoc et~al.(2020)Nguyen-Phuoc, Richardt, Mai, Yang and Mitra}]{nguyen2020blockgan}
\bibinfo{author}{Nguyen-Phuoc\xfnm[ TH]}, \bibinfo{author}{Richardt\xfnm[ C]}, \bibinfo{author}{Mai\xfnm[ L]}, \bibinfo{author}{Yang\xfnm[ Y]}, \bibinfo{author}{Mitra\xfnm[ N]}.
\newblock \bibinfo{title}{Blockgan: Learning 3d object-aware scene representations from unlabelled images}.
\newblock \bibinfo{journal}{Advances in Neural Information Processing Systems} \bibinfo{year}{2020};\bibinfo{volume}{33}:\bibinfo{pages}{6767--6778}.
\bibitem[{Chan et~al.(2021)Chan, Monteiro, Kellnhofer, Wu and Wetzstein}]{chan2021pi}
\bibinfo{author}{Chan\xfnm[ ER]}, \bibinfo{author}{Monteiro\xfnm[ M]}, \bibinfo{author}{Kellnhofer\xfnm[ P]}, \bibinfo{author}{Wu\xfnm[ J]}, \bibinfo{author}{Wetzstein\xfnm[ G]}.
\newblock \bibinfo{title}{pi-gan: Periodic implicit generative adversarial networks for 3d-aware image synthesis}.
\newblock In: \bibinfo{booktitle}{Proceedings of the IEEE/CVF conference on computer vision and pattern recognition}. \bibinfo{year}{2021}, p. \bibinfo{pages}{5799--5809}.
\bibitem[{Pan et~al.(2021)Pan, Xu, Loy, Theobalt and Dai}]{pan2021shading}
\bibinfo{author}{Pan\xfnm[ X]}, \bibinfo{author}{Xu\xfnm[ X]}, \bibinfo{author}{Loy\xfnm[ CC]}, \bibinfo{author}{Theobalt\xfnm[ C]}, \bibinfo{author}{Dai\xfnm[ B]}.
\newblock \bibinfo{title}{A shading-guided generative implicit model for shape-accurate 3d-aware image synthesis}.
\newblock \bibinfo{journal}{Advances in Neural Information Processing Systems} \bibinfo{year}{2021};\bibinfo{volume}{34}:\bibinfo{pages}{20002--20013}.
\bibitem[{Zhou et~al.(2021)Zhou, Xie, Ni and Tian}]{zhou2021cips}
\bibinfo{author}{Zhou\xfnm[ P]}, \bibinfo{author}{Xie\xfnm[ L]}, \bibinfo{author}{Ni\xfnm[ B]}, \bibinfo{author}{Tian\xfnm[ Q]}.
\newblock \bibinfo{title}{Cips-3d: A 3d-aware generator of gans based on conditionally-independent pixel synthesis}.
\newblock \bibinfo{journal}{arXiv preprint arXiv:211009788} \bibinfo{year}{2021};.
\bibitem[{Xiang et~al.(2022)Xiang, Yang, Deng and Tong}]{xiang2022gram}
\bibinfo{author}{Xiang\xfnm[ J]}, \bibinfo{author}{Yang\xfnm[ J]}, \bibinfo{author}{Deng\xfnm[ Y]}, \bibinfo{author}{Tong\xfnm[ X]}.
\newblock \bibinfo{title}{Gram-hd: 3d-consistent image generation at high resolution with generative radiance manifolds}.
\newblock \bibinfo{journal}{arXiv preprint arXiv:220607255} \bibinfo{year}{2022};.
\bibitem[{Kim et~al.(2021)Kim, Choi, Kim, Yoo and Uh}]{kim2021exploiting}
\bibinfo{author}{Kim\xfnm[ H]}, \bibinfo{author}{Choi\xfnm[ Y]}, \bibinfo{author}{Kim\xfnm[ J]}, \bibinfo{author}{Yoo\xfnm[ S]}, \bibinfo{author}{Uh\xfnm[ Y]}.
\newblock \bibinfo{title}{Exploiting spatial dimensions of latent in gan for real-time image editing}.
\newblock In: \bibinfo{booktitle}{Proceedings of the IEEE/CVF Conference on Computer Vision and Pattern Recognition}. \bibinfo{year}{2021}, p. \bibinfo{pages}{852--861}.
\bibitem[{Kwon and Ye(2021)}]{kwon2021diagonal}
\bibinfo{author}{Kwon\xfnm[ G]}, \bibinfo{author}{Ye\xfnm[ JC]}.
\newblock \bibinfo{title}{Diagonal attention and style-based gan for content-style disentanglement in image generation and translation}.
\newblock In: \bibinfo{booktitle}{Proceedings of the IEEE/CVF International Conference on Computer Vision}. \bibinfo{year}{2021}, p. \bibinfo{pages}{13980--13989}.
\bibitem[{Xu et~al.(2022{\natexlab{b}})Xu, Yin, Jiang, Wu, Zheng, Loy et~al.}]{xu2022transeditor}
\bibinfo{author}{Xu\xfnm[ Y]}, \bibinfo{author}{Yin\xfnm[ Y]}, \bibinfo{author}{Jiang\xfnm[ L]}, \bibinfo{author}{Wu\xfnm[ Q]}, \bibinfo{author}{Zheng\xfnm[ C]}, \bibinfo{author}{Loy\xfnm[ CC]}, et~al.
\newblock \bibinfo{title}{Transeditor: transformer-based dual-space gan for highly controllable facial editing}.
\newblock In: \bibinfo{booktitle}{Proceedings of the IEEE/CVF Conference on Computer Vision and Pattern Recognition}. \bibinfo{year}{2022}{\natexlab{b}}, p. \bibinfo{pages}{7683--7692}.
\bibitem[{Kafri et~al.(2021)Kafri, Patashnik, Alaluf and Cohen-Or}]{kafri2021stylefusion}
\bibinfo{author}{Kafri\xfnm[ O]}, \bibinfo{author}{Patashnik\xfnm[ O]}, \bibinfo{author}{Alaluf\xfnm[ Y]}, \bibinfo{author}{Cohen-Or\xfnm[ D]}.
\newblock \bibinfo{title}{Stylefusion: A generative model for disentangling spatial segments}.
\newblock \bibinfo{journal}{arXiv preprint arXiv:210707437} \bibinfo{year}{2021};.
\bibitem[{Sun et~al.(2022{\natexlab{b}})Sun, Wang, Zhang, Li, Zhang, Liu et~al.}]{sun2022fenerf}
\bibinfo{author}{Sun\xfnm[ J]}, \bibinfo{author}{Wang\xfnm[ X]}, \bibinfo{author}{Zhang\xfnm[ Y]}, \bibinfo{author}{Li\xfnm[ X]}, \bibinfo{author}{Zhang\xfnm[ Q]}, \bibinfo{author}{Liu\xfnm[ Y]}, et~al.
\newblock \bibinfo{title}{Fenerf: Face editing in neural radiance fields}.
\newblock In: \bibinfo{booktitle}{Proceedings of the IEEE/CVF Conference on Computer Vision and Pattern Recognition}. \bibinfo{year}{2022}{\natexlab{b}}, p. \bibinfo{pages}{7672--7682}.
\bibitem[{Chen et~al.(2022)Chen, Wu, Zheng, Cham and Cai}]{chen2022sem2nerf}
\bibinfo{author}{Chen\xfnm[ Y]}, \bibinfo{author}{Wu\xfnm[ Q]}, \bibinfo{author}{Zheng\xfnm[ C]}, \bibinfo{author}{Cham\xfnm[ TJ]}, \bibinfo{author}{Cai\xfnm[ J]}.
\newblock \bibinfo{title}{Sem2nerf: Converting single-view semantic masks to neural radiance fields}.
\newblock \bibinfo{journal}{arXiv preprint arXiv:220310821} \bibinfo{year}{2022};.
\bibitem[{Zhang et~al.(2022)Zhang, Siarohin, Liu, Tang, Sebe and Wang}]{zhang2022training}
\bibinfo{author}{Zhang\xfnm[ J]}, \bibinfo{author}{Siarohin\xfnm[ A]}, \bibinfo{author}{Liu\xfnm[ Y]}, \bibinfo{author}{Tang\xfnm[ H]}, \bibinfo{author}{Sebe\xfnm[ N]}, \bibinfo{author}{Wang\xfnm[ W]}.
\newblock \bibinfo{title}{Training and tuning generative neural radiance fields for attribute-conditional 3d-aware face generation}.
\newblock \bibinfo{journal}{arXiv preprint arXiv:220812550} \bibinfo{year}{2022};.
\bibitem[{Azadi et~al.(2020)Azadi, Pathak, Ebrahimi and Darrell}]{azadi2020compositional}
\bibinfo{author}{Azadi\xfnm[ S]}, \bibinfo{author}{Pathak\xfnm[ D]}, \bibinfo{author}{Ebrahimi\xfnm[ S]}, \bibinfo{author}{Darrell\xfnm[ T]}.
\newblock \bibinfo{title}{Compositional gan: Learning image-conditional binary composition}.
\newblock \bibinfo{journal}{International Journal of Computer Vision} \bibinfo{year}{2020};\bibinfo{volume}{128}:\bibinfo{pages}{2570--2585}.
\bibitem[{Sbai et~al.(2021)Sbai, Couprie and Aubry}]{sbai2021surprising}
\bibinfo{author}{Sbai\xfnm[ O]}, \bibinfo{author}{Couprie\xfnm[ C]}, \bibinfo{author}{Aubry\xfnm[ M]}.
\newblock \bibinfo{title}{Surprising image compositions}.
\newblock In: \bibinfo{booktitle}{Proceedings of the IEEE/CVF Conference on Computer Vision and Pattern Recognition}. \bibinfo{year}{2021}, p. \bibinfo{pages}{3926--3930}.
\bibitem[{Zhou et~al.(2022)Zhou, Ma, Zhang, Gao, Mahdavi-Amiri and Zhang}]{zhou2022}
\bibinfo{author}{Zhou\xfnm[ H]}, \bibinfo{author}{Ma\xfnm[ R]}, \bibinfo{author}{Zhang\xfnm[ LX]}, \bibinfo{author}{Gao\xfnm[ L]}, \bibinfo{author}{Mahdavi-Amiri\xfnm[ A]}, \bibinfo{author}{Zhang\xfnm[ H]}.
\newblock \bibinfo{title}{Sac-gan: Structure-aware image composition}.
\newblock \bibinfo{journal}{IEEE Transactions on Visualization and Computer Graphics} \bibinfo{year}{2022};:\bibinfo{pages}{1--13}.
\bibitem[{Eslami et~al.(2016)Eslami, Heess, Weber, Tassa, Szepesvari, Hinton et~al.}]{eslami2016attend}
\bibinfo{author}{Eslami\xfnm[ S]}, \bibinfo{author}{Heess\xfnm[ N]}, \bibinfo{author}{Weber\xfnm[ T]}, \bibinfo{author}{Tassa\xfnm[ Y]}, \bibinfo{author}{Szepesvari\xfnm[ D]}, \bibinfo{author}{Hinton\xfnm[ GE]}, et~al.
\newblock \bibinfo{title}{Attend, infer, repeat: Fast scene understanding with generative models}.
\newblock \bibinfo{journal}{Advances in neural information processing systems} \bibinfo{year}{2016};\bibinfo{volume}{29}.
\bibitem[{Yang et~al.(2017)Yang, Kannan, Batra and Parikh}]{yang2017lr}
\bibinfo{author}{Yang\xfnm[ J]}, \bibinfo{author}{Kannan\xfnm[ A]}, \bibinfo{author}{Batra\xfnm[ D]}, \bibinfo{author}{Parikh\xfnm[ D]}.
\newblock \bibinfo{title}{Lr-gan: Layered recursive generative adversarial networks for image generation}.
\newblock \bibinfo{journal}{arXiv preprint arXiv:170301560} \bibinfo{year}{2017};.
\bibitem[{Burgess et~al.(2019)Burgess, Matthey, Watters, Kabra, Higgins, Botvinick et~al.}]{burgess2019monet}
\bibinfo{author}{Burgess\xfnm[ CP]}, \bibinfo{author}{Matthey\xfnm[ L]}, \bibinfo{author}{Watters\xfnm[ N]}, \bibinfo{author}{Kabra\xfnm[ R]}, \bibinfo{author}{Higgins\xfnm[ I]}, \bibinfo{author}{Botvinick\xfnm[ M]}, et~al.
\newblock \bibinfo{title}{Monet: Unsupervised scene decomposition and representation}.
\newblock \bibinfo{journal}{arXiv preprint arXiv:190111390} \bibinfo{year}{2019};.
\bibitem[{Greff et~al.(2019)Greff, Kaufman, Kabra, Watters, Burgess, Zoran et~al.}]{greff2019multi}
\bibinfo{author}{Greff\xfnm[ K]}, \bibinfo{author}{Kaufman\xfnm[ RL]}, \bibinfo{author}{Kabra\xfnm[ R]}, \bibinfo{author}{Watters\xfnm[ N]}, \bibinfo{author}{Burgess\xfnm[ C]}, \bibinfo{author}{Zoran\xfnm[ D]}, et~al.
\newblock \bibinfo{title}{Multi-object representation learning with iterative variational inference}.
\newblock In: \bibinfo{booktitle}{International Conference on Machine Learning}. \bibinfo{organization}{PMLR}; \bibinfo{year}{2019}, p. \bibinfo{pages}{2424--2433}.
\bibitem[{Ehrhardt et~al.(2020)Ehrhardt, Groth, Monszpart, Engelcke, Posner, Mitra et~al.}]{ehrhardt2020relate}
\bibinfo{author}{Ehrhardt\xfnm[ S]}, \bibinfo{author}{Groth\xfnm[ O]}, \bibinfo{author}{Monszpart\xfnm[ A]}, \bibinfo{author}{Engelcke\xfnm[ M]}, \bibinfo{author}{Posner\xfnm[ I]}, \bibinfo{author}{Mitra\xfnm[ N]}, et~al.
\newblock \bibinfo{title}{Relate: Physically plausible multi-object scene synthesis using structured latent spaces}.
\newblock \bibinfo{journal}{Advances in Neural Information Processing Systems} \bibinfo{year}{2020};\bibinfo{volume}{33}:\bibinfo{pages}{11202--11213}.
\bibitem[{Arad~Hudson and Zitnick(2021)}]{arad2021compositional}
\bibinfo{author}{Arad~Hudson\xfnm[ D]}, \bibinfo{author}{Zitnick\xfnm[ L]}.
\newblock \bibinfo{title}{Compositional transformers for scene generation}.
\newblock \bibinfo{journal}{Advances in Neural Information Processing Systems} \bibinfo{year}{2021};\bibinfo{volume}{34}:\bibinfo{pages}{9506--9520}.
\bibitem[{Niemeyer and Geiger(2021)}]{niemeyer2021giraffe}
\bibinfo{author}{Niemeyer\xfnm[ M]}, \bibinfo{author}{Geiger\xfnm[ A]}.
\newblock \bibinfo{title}{Giraffe: Representing scenes as compositional generative neural feature fields}.
\newblock In: \bibinfo{booktitle}{Proceedings of the IEEE/CVF Conference on Computer Vision and Pattern Recognition}. \bibinfo{year}{2021}, p. \bibinfo{pages}{11453--11464}.
\bibitem[{Xue et~al.(2022)Xue, Li, Singh and Lee}]{xue2022giraffe}
\bibinfo{author}{Xue\xfnm[ Y]}, \bibinfo{author}{Li\xfnm[ Y]}, \bibinfo{author}{Singh\xfnm[ KK]}, \bibinfo{author}{Lee\xfnm[ YJ]}.
\newblock \bibinfo{title}{Giraffe hd: A high-resolution 3d-aware generative model}.
\newblock In: \bibinfo{booktitle}{Proceedings of the IEEE/CVF Conference on Computer Vision and Pattern Recognition}. \bibinfo{year}{2022}, p. \bibinfo{pages}{18440--18449}.
\bibitem[{Lee et~al.(2020)Lee, Liu, Wu and Luo}]{lee2020maskgan}
\bibinfo{author}{Lee\xfnm[ CH]}, \bibinfo{author}{Liu\xfnm[ Z]}, \bibinfo{author}{Wu\xfnm[ L]}, \bibinfo{author}{Luo\xfnm[ P]}.
\newblock \bibinfo{title}{Maskgan: Towards diverse and interactive facial image manipulation}.
\newblock In: \bibinfo{booktitle}{Proceedings of the IEEE/CVF Conference on Computer Vision and Pattern Recognition}. \bibinfo{year}{2020}, p. \bibinfo{pages}{5549--5558}.
\bibitem[{Heusel et~al.(2017)Heusel, Ramsauer, Unterthiner, Nessler and Hochreiter}]{heusel2017gans}
\bibinfo{author}{Heusel\xfnm[ M]}, \bibinfo{author}{Ramsauer\xfnm[ H]}, \bibinfo{author}{Unterthiner\xfnm[ T]}, \bibinfo{author}{Nessler\xfnm[ B]}, \bibinfo{author}{Hochreiter\xfnm[ S]}.
\newblock \bibinfo{title}{Gans trained by a two time-scale update rule converge to a local nash equilibrium}.
\newblock \bibinfo{journal}{Advances in neural information processing systems} \bibinfo{year}{2017};\bibinfo{volume}{30}.

\end{thebibliography}

\end{document}